\documentclass{article}

\usepackage{graphicx} % Required for inserting images
\usepackage{float}
\usepackage{authblk}
\usepackage{url}

\usepackage{algorithm}
\usepackage{algorithmic}
\usepackage{multirow}
\usepackage{booktabs}
\usepackage{float}
\usepackage{makecell}
\usepackage{makecell}
\usepackage{CJKutf8}
\usepackage[numbers,sort&compress]{natbib}
\usepackage{enumitem}

\usepackage[colorlinks=true, urlcolor=blue, linkcolor=red]{hyperref}

\title{LLMs for Coding and Robotics Education}

\author{Peng Shu, Huaqin Zhao, Hanqi Jiang, Yiwei Li, Shaochen Xu, Yi Pan, Zihao Wu, Zhengliang Liu, Guoyu Lu, Le Guan, Gong Chen, Xianqiao Wang Tianming Liu$\dagger$\thanks{$\dagger$Corresponding authors: Tianming Liu}\thanks{ Peng Shu, Huaqin Zhao, Hanqi Jiang, Yiwei Li, Shaochen Xu, Yi Pan, Zihao Wu, Zhengliang Liu, Guoyu Lu, Le Guan, Gong Chen, Tianming Liu are with the School of Computing, The University of Georgia, Athens 30602, USA. Xianqiao Wang is with School of ECAM, College of Engineering, The University of Georgia, Athens 30602, USA.}}

\begin{document}
% \begin{CJK}{UTF8}{gbsn}

\maketitle

\begin{abstract}
Large language models and multimodal large language models have revolutionized artificial intelligence recently. An increasing number of regions are now embracing these advanced technologies. Within this context, robot coding education is garnering increasing attention. To teach young children how to code and compete in robot challenges, large language models are being utilized for robot code explanation, generation, and modification. In this paper, we highlight an important trend in robot coding education. We test several mainstream large language models on both traditional coding tasks and the more challenging task of robot code generation, which includes block diagrams. Our results show that GPT-4V outperforms other models in all of our tests but struggles with generating block diagram images.
\end{abstract}

\section{Introduction}
The advent of Large Language Models (LLMs) and multimodal foundation models (FMs) such as  GPT-4V\cite{achiam2023gpt}, BERT\cite{devlin2018bert}, Llama2\cite{touvron2023llama}, marks a significant milestone in the field of artificial general intelligence (AGI) and natural language processing (NLP). These models are usually trained on vast datasets with enormous amount of samples, which enables them with advanced generalization. Basically, they can understand and generate human-like text. These transformer\cite{vaswani2017attention} structure models can also apply superior capabilities in zero-shot or few-shots learning for the downstream tasks\cite{liu2023summary}\cite{zhao2023brain}\cite{liu2023visual}. LLMs and FMs have raised a wide and profound revolution in industries by challenging traditional working patterns. Among these area, coding is one of the particular regions that being affected most.

The influence of LLMs on coding is multifaceted. Firstly, it revolutionized code generation and understanding. Based on their generalization into the domain of coding, LLMs serve as powerful tools for code completion, automated documentation generation, and intelligent debugging. Tools such as GitHub's Copilot\cite{kalliamvakou2022research}, powered by OpenAI's Codex, offer suggestions to improve code quality and efficiency, making coding more accessible to a wider audience. With help of AI assistants, programmers get rid of duplicate code work, refactor with higher efficiency and more readable code. Secondly, LLMs enhance code development process productivity and efficiency. LLMs serve as tireless partners in the coding process, available around the clock to assist developers. They automate routine tasks, suggest optimizations, and even identify potential bugs before they become problematic, which significantly improves code quality and efficiency\cite{poldrack2023ai}. Thirdly, LLMs play a virtal role in democratizing coding and coding education. By providing intuitive suggestions and reducing the need for intricate understanding of syntax and language-specific idiosyncrasies, LLMs lower the barrier to entry for coding. This has opened up the field of software development to a broader demographic, fostering a more inclusive and diverse tech community. However, given that the current
models can easily complete most coding problem sets using in introductory programming courses\cite{finnie2022robots}, there are more and more concerns about necessity of basic education in code programming and the unemployment problem of programmers.

In recent years there have been different attempts to incorporate LLMs into robotics systems. \cite{vemprala2023chatgpt} provides a clear and concise description of the desired robotics task and its context to generate more accurate responses. \cite{wang2024large} applies GPT-4V to generate robotic code commands given the tasks and vision input. Other works put forward similar robotic framework cooperated with LLMs\cite{jin2024robotgpt}\cite{finnie2022robots}. LLMs make it possible for machines to interpret and act on complex instructions and queries, which significantly changes the way humans interact with robots, shifting from rigid command-based communication to more natural and intuitive dialogues. Robots can now understand context, process ambiguous inputs, and even engage in conversational exchanges, making them more accessible and user-friendly. In robot coding, LLMs contribute to more autonomous systems capable of decision-making based on linguistic cues and context. This autonomy is particularly impactful in scenarios where robots need to understand and adapt to dynamic environments, such as in search and rescue operations, healthcare, and customer service. However, the realm in understanding robotic code and generating robot function modular code remains almost blank, especially for early programming education. In this paper, we demonstrate the performance of LLMs on regular data structure and algorithms code testing as well as robot module code generation, understanding of block diagram robot languages. We also test and attempt to generate fundamental block diagrams for robot education. Discussion includes how LLMs and FMs can support robot education for children, the possible directions for future robot programming education based on LLMs and current limitations LLMs have in this region.
\section{Related Work}

\subsection{Large Language Models}
The introduction of GPT-3 by Brown et al.\cite{brown2020language} provided an innovative basis for subsequent LLM applications in coding, demonstrating the model's ability to generate human-like text, including code. This work laid the groundwork for specialized models like Codex\cite{chen2021evaluating}, further tailored for programming languages and code generation tasks. Similarly, Devlin et al.\cite{devlin2018bert} introduced BERT, which, although not specifically designed for code, has inspired successive adaptations like CodeBERT by Feng et al.\cite{feng2020codebert}, merging natural language understanding with code semantics.

Following the path blazed by Codex, Alon et al.\cite{alon2019code2seq} proposed the Code2Seq framework, which represents a novel approach to generating code summaries and assisting in code documentation by converting code into a sequence of tokens. This model emphasizes the importance of understanding the structural and syntactical implications of programming languages in LLM applications. Additionally, Lu et al.\cite{lu2021codexglue} introduced CodeGPT, a variation of GPT specifically fine-tuned for code generation tasks, showcasing improvements in generating syntactically correct and logically coherent code snippets.

LLMs, such as OpenAI's Codex\cite{chen2021evaluating}, which powers GitHub Copilot, have significantly enhanced developer productivity by providing code suggestions and completions. Codex, built on the GPT-3 model, demonstrates an advanced ability to generate programming code based on natural language prompts, thereby accelerating the software development process and reducing manual coding effort. Research on utilizing LLMs for debugging and error correction has shown promising results. For instance, a model like CodeBERT\cite{feng2020codebert}, designed to understand and generate code from natural language, can be applied to identify syntax errors and logical bugs within software projects, providing a foundation for automated debugging tools. The security of code generated by LLMs and the mitigation of biases within these models are critical concerns. Research efforts such as those by Hajipour et al.\cite{hajipour2023codelmsec} have focused on developing methodologies to ensure the generation of secure code and to identify potential biases in model outputs, aiming to establish best practices for the safe and ethical use of LLMs in software development.

%\subsection{LLMs in Code Generation and Bug Detection}
\subsection{Robot Coding for Education}
In recent years, numerous organizations have dedicated efforts to identify essential life skills for students in educational environment, which inspires plenty of researches in this area. As a result, these skills are summarized as 21st century skills, and a basic frame for these skills has been attempted to be designed \cite{ccakir2021effect}. Organisation for Economic Co-operation and Development (OECD) attempt to divide the skills into three categories: Creating new value (being able to develop new perspectives), reconciling tensions and dilemmas (being able to think in a systematic way), as well as taking responsibility (self-regulation, self-efficacy, self-control, and problem-solving)\cite{taguma2019oecd}.

Among required skills, Learning and Innovation Skills (creativity and innovation, critical
thinking and problem-solving, communication, collaboration) are regraded as one of the most important realms for development of teenagers. They promise the success of students in such a information explosion century by reaching and filtering valuable information efficiently, innovating on current platform and cooperating within a team. Problem-solving involves children's understanding on concepts, discovering the concepts and values independently and developing them\cite{siahaan2017improving}. To empower young people for problem-solving, previous methods include structured or unstructured gaming
activities, brainstorming, project-based education and etc\cite{ccakir2021effect,yiugitalp2014yonlendirilmics,ouguz2012proje}. Recently, coding has become a systematic method in terms of solving problems, which prompt teenagers to improve critical thinking, computational thinking and innovative thinking\cite{bocconi2016developing,kanaki2018introducing,odaci2017okul}. More robotics and coding instruction, commonly preferred in primary and
secondary education, has been used as a teaching technique in preschool education\cite{ccakir2021effect}. 
\cite{chun2020effect} observed that after a brief period of five weeks engaged in unplugged coding activities, a group of 5-year-old children demonstrated marked progress in the development of creative and social personality traits. Other researches show that teaching basic practices for robot and coding supports sequencing skills in children\cite{caballero2019learning,kazakoff2014put,saxena2020designing}. Teenagers who take part in robotics and coding activities outperform in skills vary from visual-spatial, working memory, inhibitory control to problem-solving\cite{akyol2018okul,kocc2019okul,di2017educational}.

However, coding is an abstract and complicated method because coding itself can be defined as a problem-solving procedure. Robotics combines software code design as well as robotic structure, which requires even higher comprehensive capability. The difficulty of coding and robotics becomes the barrier hindering young children from learning. For this reason, numerous educational platforms have been developed to facilitate learning in coding and robotics. Such platforms include WeDo, Lego, Bee-bot, Clementino, and Robokids. Generally, these platforms are equipped with coding kits and designed with a visual interface, making them more user-friendly and easier for young children to learn. Children consider them easy and funny to learn, thus promoting their passion.

Other challenges lie on how to provide concrete, meaningful and problem-based learning activities to teach coding to young children\cite{dorouka2020tablets}. \cite{bers2020coding} demonstrates the importance of proper coding and robotics activities. It also emphasizes the participation of children in terms of playing with robots, exploring unknown regions , socializing with others and creating innovative solutions\cite{lee2020coding}. Besides, coding and robotics education is a new concept for early childhood teachers and has not been comprehensively integrated into most early childhood curricula\cite{canbeldek2023exploring}. All of these existing problems motivate this paper to investigate the possibility of applying LLMs in robotics coding to provide assistance to both children and teachers.
\subsection{LLMs in Robot Code Generation}

\subsubsection{FIRST Tech Challenge}
The FIRST Tech Challenge (FTC) is a dynamic, global robotics program that ignites passion in young minds for science, technology, engineering, and mathematics (STEM). The FTC robot code is typically written in Java. Established in 2005, FTC provides a platform for students in grades 4-18 to engage in hands-on robotics challenges, fostering invaluable skills in problem-solving, teamwork, and innovation \cite{ftcabout2024}. By designing, building, and programming robots, participants are immersed in real-world engineering experiences, competing in alliances against teams at local, regional, and international levels \cite{ftcabout2024}.

The program is a flagship initiative of For Inspiration and Recognition of Science and Technology (FIRST), a non-profit organization founded by inventor Dean Kamen in 1989. FIRST's mission is to inspire young people to become leaders in science and technology. FTC plays a crucial role in providing a platform which is accessible and inclusive, regardless of a participant's background or experience level \cite{ftcabout2024}.

The impact of FTC extends beyond the technical skills. Research indicates that participants are significantly more likely to attend college and major in a STEM field, demonstrating the program's effectiveness in shaping the next generation of STEM professionals \cite{ftcimpact2023}. Moreover, FTC's emphasis on teamwork, communication, and leadership prepares students for future challenges, aligning with educational goals around the globe \cite{iturbe2009educating}.

\subsubsection{FIRST LEGO League}
FIRST LEGO League (FLL) is an internationally recognized, innovative program that ignites enthusiasm for discovery, science, and technology in young minds. Designed to inspire and challenge students aged 4 to 16, the league combines the hands-on fun of LEGO building with real-world engineering and problem-solving challenges\cite{fllabout2024}. Originating from a partnership between FIRST and the LEGO Group, FLL provides a platform for children to learn critical skills while engaging in playful and meaningful competition.

The structure of FLL is centered around theme-based Challenges that teams must navigate through. In these Challenges, participants build and program autonomous robots using LEGO MINDSTORMS technology to score points on a thematic playing surface, creating innovative solutions to a problem as part of their research project. The process fosters valuable life skills and competencies such as problem-solving, teamwork, and creative thinking.

FLL block diagrams are a fundamental component of the FLL robotics experience, offering a visual and intuitive way to program LEGO MINDSTORMS robots. These diagrams are designed to be user-friendly, ensuring that even individuals without prior programming experience can engage with and understand the basics of robot programming.
The essence of FLL block diagrams lies in their drag-and-drop interface, where users can select from a variety of command blocks – each representing a specific action or decision in the robot's operation. These blocks can be pieced together to form a sequence of instructions, creating a 'flow' of actions that the robot will execute.

\subsubsection{Cooperating LLMs in Robot Coding}
Unlike traditional text-based coding languages like Python or Java, visual block coding platforms like Tynker, Scratch, MakeCode, and Snap operate at a higher level of abstraction. They often transcend standard programming constructs like loops and conditionals. For example, a single block like “move n pixels” could equate to 20 lines of C code. Beginners would need to navigate the intricacies of graphics libraries, variable tracking, and angular calculations to achieve the same result\cite{kelvin2023coding}. Therefore, companies are leveraging LLMs to facilitate the interaction between the robot and the young learners, allowing for more intuitive and meaningful interactions. For example, Aries Hilton’s PyScratch is a software that can convert Python files to Scratch projects, which are interactive stories, games, and animations that can be shared online. PyScratch uses LLMs to map the Python commands and arguments to the corresponding Scratch blocks and inputs, and also converts the Pygame images to SVG files. PyScratch can help Python learners to create Scratch projects without having to learn a new language, and also help Scratch users to learn Python by seeing how their projects can be translated into code\cite{aries2023pyScratch}.

\cite{kelvin2023coding} trained a model specifically designed to autocomplete the next block of code in a Tynker visual block coding project based on the BERT in 2018 using their expansive repository of millions of user-published projects. With the emergence of ChatGPT, they guide ChatGPT into generating block-like code that could be converted into Tynker’s native coding language to help kids effortlessly translate their ideas into code, understand any given code’s functionality, and provide real-time debugging assistance. Further more, they fine tune a Llama2 chat model to meet Tynker’s specialized visual coding requirements.
\section{Methodologies and Tasks}

\subsection{Data Structure and Algorithms Code Generation}

For the purpose of evaluating code generation in the context of data structures and algorithms, we selected three problems of varying difficulty levels from LeetCode to test the performance of three models: GPT-4V, Code Llama, and GitHub Copilot. Each model generates four different types of programming languages including C, C++, Python3 and Java for every problem.

The first problem, categorized as easy, involves designing a data structure that uses two stacks to implement a first-in-first-out queue. This queue should support four operations: adding an element to the end of the queue, removing and returning an elemen        t from the beginning of the queue, returning the element at the beginning of the queue, and checking whether the queue is empty. The second problem, of medium difficulty, requires determining whether a non-empty array containing only positive integers can be partitioned into two subsets with equal sums. This problem entails finding a method to group the elements in the array such that the total sum of each group is equal. The challenging problem involves determining the shortest path length to visit all nodes in an undirected connected graph composed of n nodes, where the nodes are numbered from 0 to n - 1. The input to the problem is an array, graph, representing the structure of the graph, where graph[i] is a list containing all nodes directly connected to the i-th node.

\subsection{FTC Code Generation}

In order to assess the robot code generation capabilities of GPT-4V, Code Llama, and GitHub Copilot, we selected three FTC code modules: AprilTag label recognition and pose estimation, encoder-based path driving, and operation of a four-wheel omnidirectional robot. For each of these tasks, we removed the original code but provided comments to serve as input for the model to generate code. Our objective is to evaluate the quality of code generated from the models by comparing it with the original code.

The first test code aims at recognizing and estimating the position and orientation of AprilTag tags. Operating in a LinearOpMode, it demonstrates the basics of AprilTag recognition and pose estimation, including Java Builder structures for specifying vision parameters. The essence of the program is to detect AprilTag tags in the field of view through a camera, which can be either a webcam or a built-in phone camera, and to display the IDs of these tags. For tags included in the default \textit{TagLibrary}, the program not only shows their IDs but also provides their position and orientation relative to the camera. This default TagLibrary contains the current season's AprilTags and a set of test tags in a higher number range. When an AprilTag from the TagLibrary is detected, the program acquires the tag's location and orientation relative to the camera using an instance of the AprilTag Processor (AprilTagProcessor), with this information stored in the \textit{ftcPose} member of the returned \textit{detection}. The program also includes an instance of the Vision Portal (VisionPortal) for managing the camera's streaming and processor configurations. Additionally, the code contains telemetry data logic to display the results of AprilTag detections. When the program is operational, it periodically checks and displays information for all detected AprilTag tags, including each tag's ID, position (XYZ coordinates), and orientation (pitch, roll, yaw angles).

The second test code focuses on path driving based on encoder counts. It operates in a Linear Operation Mode (LinearOpMode) and requires the installation of encoders on the robot's wheels. The core functionality of the program is executed through the defined method \textit{encoderDrive(speed, leftInches, rightInches, timeoutS)}, which assumes that each movement is relative to the robot's last stopping position. The program defines a specific path that includes driving forward for 48 inches, spinning right for 12 inches, driving backward for 24 inches, and then stopping and closing the claw. This method employs the \textit{RUN\_TO\_POSITION} mode, enabling the motor controllers to generate a run profile. The program calculates the encoder counts per inch (COUNTS\_PER\_INCH), based on the specific drive train configuration, such as motor revolutions per minute and external gear reduction ratios. During operation, the program first initializes the drive system variables, sets the direction of the motors, and resets the encoders. It then waits for the game to start (activated by the driver pressing PLAY) and subsequently executes the predefined path in steps, including forward movement, turning, and reversing. New target positions are determined through the encoder counts, and the motors are set to reach these positions. Throughout the process, the program continuously checks whether any of the termination conditions are met: reaching the desired position, running out of time, or the driver stopping the program. After each movement, the program pauses briefly and then displays the final telemetry message.

The third code segment is designed for operating a four-wheeled omnidirectional (or holonomic) robot. The program operates in a Linear Operation Mode (LinearOpMode) and is compatible with both Mecanum-Drive and X-Drive trains, which are suitable for omnidirectional wheel robots. The core functionality of the program lies in using four motors (left front, left back, right front, right back) to control the robot's three motion axes: axial (forward and backward movement), lateral (side-to-side sliding), and yaw (clockwise and counter-clockwise rotation). These motion axes are controlled by different axes of the gamepad; for instance, the left joystick controls the axial and lateral movements, while the right joystick controls rotation. A critical aspect of the program is the correct setting of the rotation direction for each motor. Instructions are provided within the code on how to test and set the motor directions. For example, if the robot moves backward when intending to move forward, the direction of all four motors needs to be reversed. Throughout the operation, the program continuously updates and displays the elapsed match time and the power level of each wheel. This information is shown through telemetry data, providing real-time feedback to the driver. The program concludes its operation (when the driver presses the STOP button) by halting all actions.

\subsection{FLL Blocks Explanation}
Beyond code generation capabilities, as demonstrated in prior evaluations, we further extended to appraise the understanding and explanation performance of the three prevailing multimodal LLMs — GPT-4V, LLaVA-1.5, and Microsoft Copilot — concerning the FLL block diagrams. These block diagrams, characterized by their intuitive drag-and-drop functionality, aim to demystify robotics programming for younger audiences, particularly those in elementary and middle schools. They facilitate the assembly of command sequences essential for controlling robot movement and sensor responses. To assess the LLMs' comprehension of FLL block diagrams, three specific tasks were delineated: improving pivot turn accuracy, controlling motor movement with color sensor, and adjusting the speed of the motors based on the force sensor. Each task involves presenting the FLL block diagrams as inputs to the aforementioned LLMs, which are then tasked with articulating the purpose and function of these blocks.

The initial task is aimed at enhancing the precision of pivot turns within robotics applications. The selected FLL blocks propose strategies to augment the accuracy of pivot turns, acknowledging potential overshoots in turn angles due to gyroscopic read delays and inherent robot momentum (e.g., a robot executing a 102-degree turn instead of the intended 90 degrees). The blocks recommend adjustments in the turn angles to offset these discrepancies, factoring in the robot's velocity and design specifics. An example provided is the Droid Bot IV configuration, which suggests a 12-degree correction to secure an exact 90-degree turn.

The second task introduces the integration of color sensor within robotics programming, employing Wait Until Blocks for enhanced control. The FLL blocks tested offer a strategy for programming a robot to proceed until it detects black regions with its color sensor. This approach involves setting up the movement motors and speed, initiating the robot's movement, using the Wait Until Block to await the detection of black regions, and subsequently halting the robot. This scenario exemplifies the practical application of sensor inputs in robotic programming for accomplishing specific objectives.

Finally, the third task focuses on the application of Repeat Blocks for dynamic motor control, based on force-sensory feedback. The FLL programming blocks provided relate to adjusting a robot's motor speed dynamically, contingent upon the force detected by a sensor. This loop persists until the force sensor's activation ceases, illustrating the utility of these blocks in executing operations pending the fulfillment of a specified condition. This guidance is pivotal for cultivating foundational robotics skills that enable the development of responsive and adaptable robot behaviors.

\subsection{Visual Programming Code Generation}
In the realm of Visual Programming Code Generation, research has explored innovative approaches to translating visual inputs into executable code. Among these approaches, leveraging multimodal models to interpret and generate code from visual diagrams, such as the FIRST LEGO League (FLL) block diagrams, represents a cutting-edge frontier. ChainCoder introduces a multi-pass code generation framework that uses a unique syntax hierarchy to enhance step-wise reasoning in LLMs, aiming to improve code accuracy and syntactic coherence. It employs a syntax-aware tokenizer and a specialized transformer to leverage structured, syntactically aligned data for progressive, multi-level code generation\cite{zheng2023outline}. VISPROG is a system that leverages the in-context learning capabilities of language models to transform natural language instructions into visual programs for complex visual tasks, enabling the creation of sophisticated visual solutions directly from user inputs\cite{gupta2023visual}. However, our investigations reveal that not all multimodal models are equally proficient at generating high-quality code from such visual inputs. For instance, while GPT-4V demonstrates remarkable capability in this domain, another model, Copilot, falls short. This discrepancy may stem from several factors. First, Copilot might struggle with prompts not written in English, highlighting a potential language bias. Second, its generation and evaluation modules, primarily trained on natural images, may not effectively interpret visual diagrams, which are fundamentally different in structure and content from typical photographic imagery.

In light of these challenges, our research pivoted towards leveraging textual prompts to generate pseudocode as an intermediary step. This approach capitalizes on the superior textual processing capabilities of multimodal models over their visual processing counterparts. By instructing the model to generate step-by-step, finely represented pseudocode instead of a coarse overview, we harness the model's linguistic strengths. Our prompts are meticulously crafted to guide the model in this process, resulting in more accurate and functional code generation.

Our testing, conducted in a zero-shot learning context, underscores the potential of these models. However, it also opens avenues for further exploration into few-shot learning and fine-tuning strategies. Such enhancements could significantly improve the models' code generation capabilities, especially when dealing with complex visual inputs or domain-specific requirements. As we continue to refine our methodologies, the anticipation grows for what future iterations of these large models might achieve in the field of Visual Programming Code Generation, blending the visual and textual realms to automate and innovate in coding practices.

\section{Experimental Results}

Table \ref{Leetcode test} shows the testing results for Leetcode problems.
\begin{table}[ht]
\centering
\begin{tabular}{cccc}
\hline
Models        & LeetCode 232 & LeetCode 416 & LeetCode 847 \\ \hline
GPT-4V          & 100\%        & 100\%        & 100\%        \\
Code Llama         & 50\%         & 50\%         & 0\%          \\
GitHub Copilot & 100\%        & 100\%        & 75\%         \\ \hline
\end{tabular}
\caption{Leetcode test pass rate for three LLMs. The percentage indicates how many programming languages the LLM generated pass the test.}
\label{Leetcode test}
\end{table}

GPT-4V passes all of three test problems for all of four programming languages while Code Llama shows the worst performance. For LeetCode 232 and LeetCode 416 corresponding to easy and medium level, only Python3 and C pass the test. The rest two of languages encounter either compile errors or wrong answer. For the hard problem, none of the answers passes the test. GitHub Copilot passes the first two problems but give wrong answer for the LeetCode 847. In this test, GPT-4V demonstrates the best performance followed by GitHub Copilot and Code Llama.

\begin{figure*}
\centering
\includegraphics[width=1\textwidth]{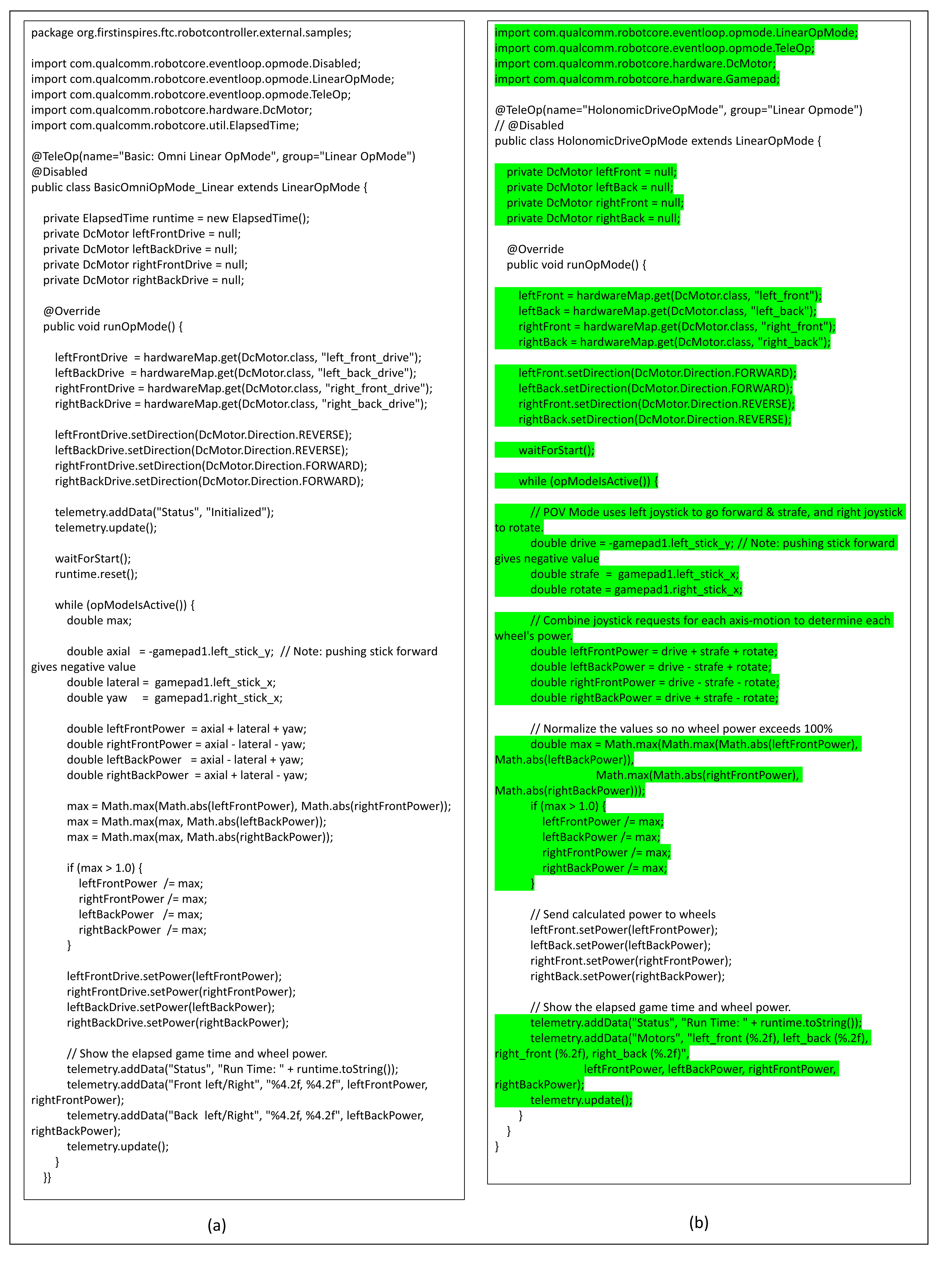}
\caption{One example for FTC code generation test results.(a) represents the original code and (b) is the code generated by GPT-4V. The green backgrounds indicate correct parts.} 
\label{gpt4_ftc}
\end{figure*}

\begin{figure*}
\centering
\includegraphics[width=1\textwidth]{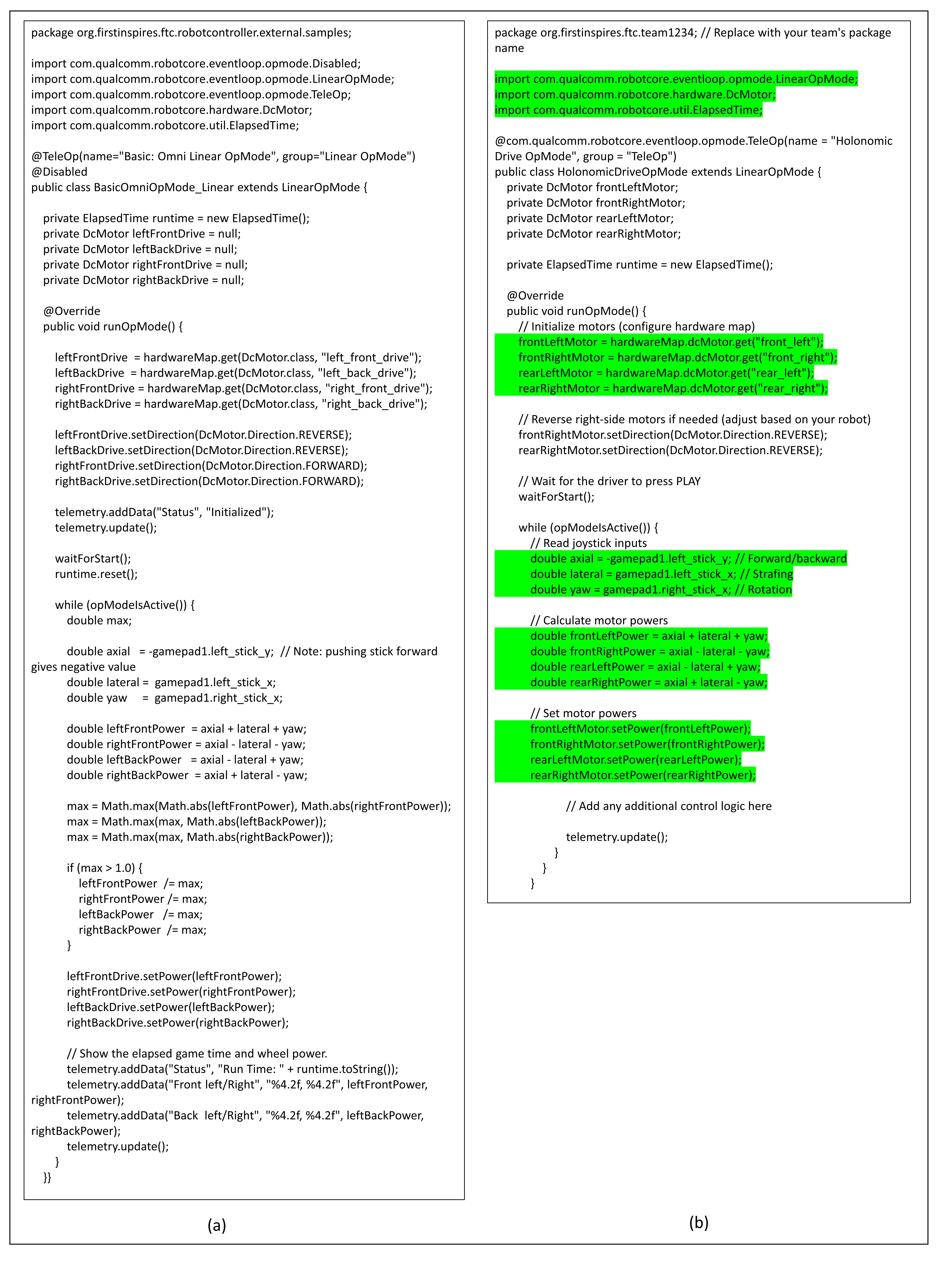}
\caption{Same example for FTC code generation test results.(a) represents the original code and (b) is the code generated by GitHub Copilot. The green backgrounds indicate correct parts.} 
\label{copilot_ftc}
\end{figure*}

Figure \ref{gpt4_ftc} and \ref{copilot_ftc} show the FTC code generation task results. Both of two models construct the complete code framework. The general logic of codes are clear with correct definitions and sequential operations. GPT-4V can recover around 90\% of the original code with minor missing parts. However, these defects lie on printed messages and extra parameters validity check, which does not have severe consequence for the main module. Compared with GPT-4V, GitHub Copilot lacks more parameters validity check. It also skips or combines some operations such as setting front and rear motor, which leads to movement failure. Besides, it does not follow the comments to show the elapsed game time and wheel power, suggesting the code is not as delicate as GPT-4V's. Code Llama fails in this task. Out test shows that Code Llama cannot understand the task. It refuses to generate code but with some explanation text. In summary, GPT-4V still exceeds other two models for FTC code generation. 

\begin{figure*}
\centering
\includegraphics[width=1\textwidth]{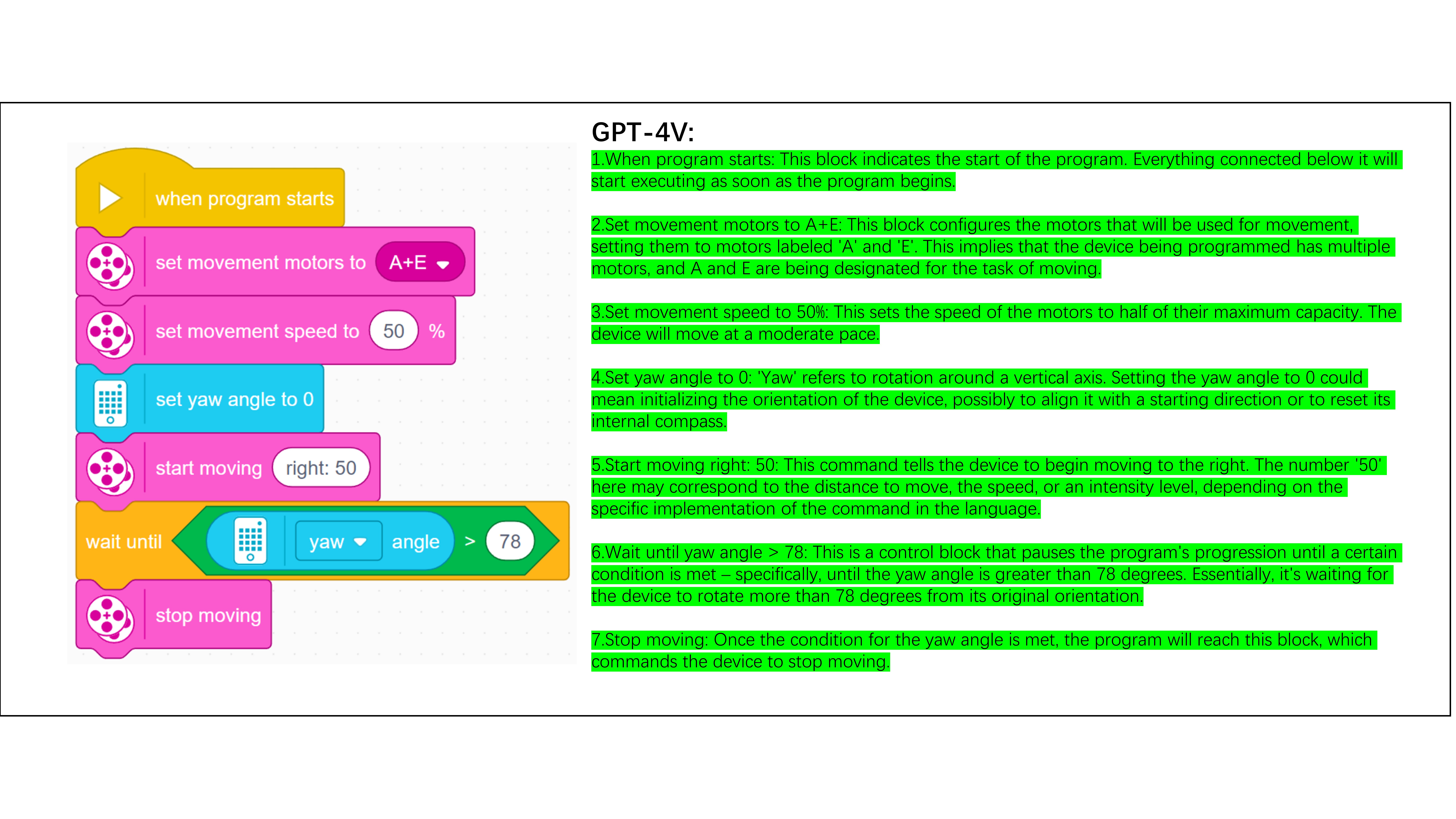}
\caption{One example for FLL block diagram explanation test results. The left code refers to the FLL block diagram while right part contains explanation from GPT-4V.} 
\label{gpt4_fll}
\end{figure*}

\begin{figure*}
\centering
\includegraphics[width=1\textwidth]{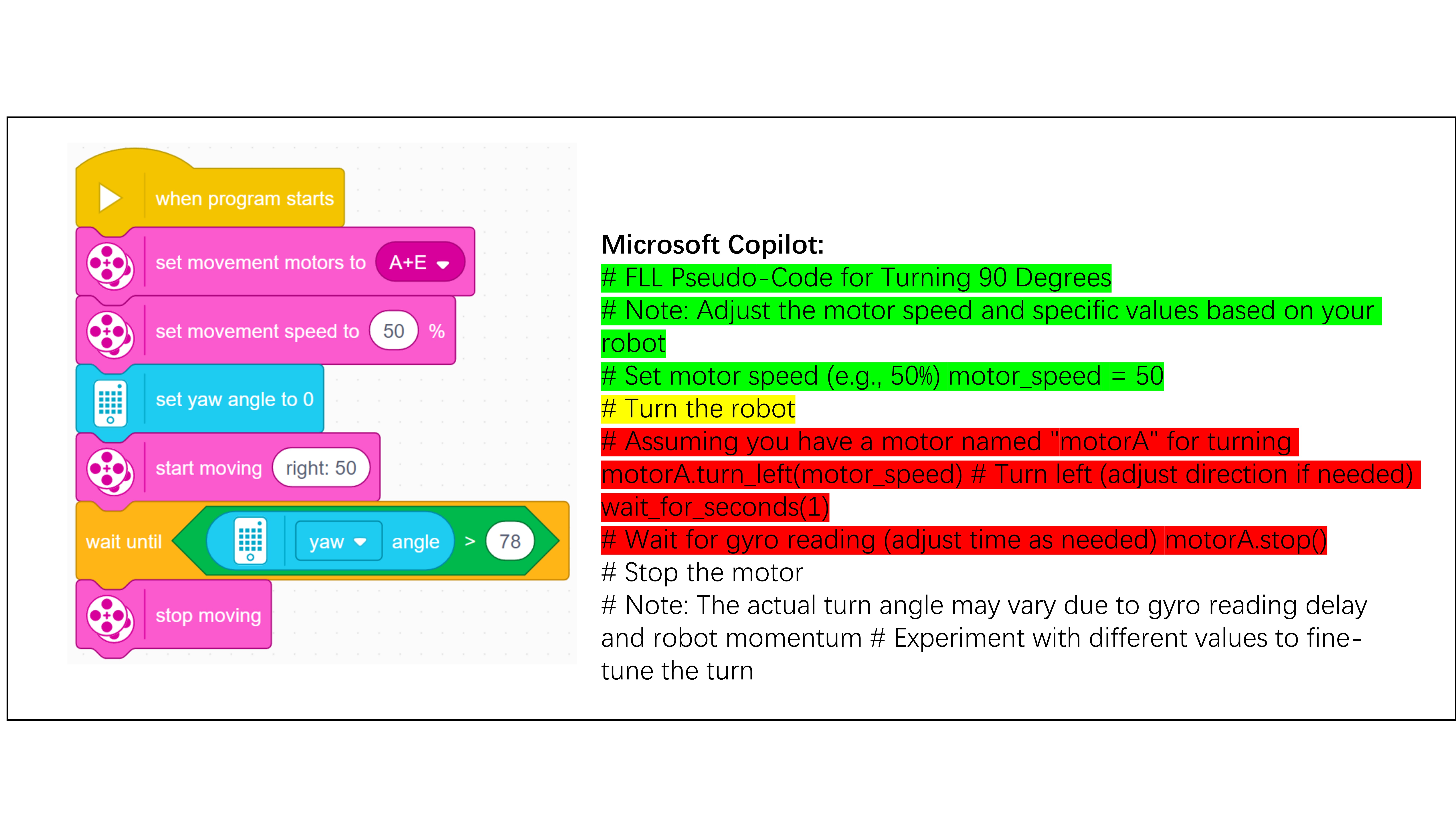}
\caption{Same example for FLL block diagram explanation test results. The left code refers to the FLL block diagram while right part contains explanation from Microsoft Copilot. The yellow background indicates inaccurate description and red for wrong steps.} 
\label{copilot_fll}
\end{figure*}

Figure \ref{gpt4_fll} and \ref{copilot_fll} demonstrate the performance for one FLL block diagram understanding and explanation. Since this task involves images analysis, we attempt three multimodal LLMs: GPT-4V, Microsoft Copilot and LLaVA-1.5. GPT-4V recognizes the diagram successfully. Although it does not specify which block diagram this image belongs to, it gives a perfect explanation for each block. The comments are accurate and clear enough so that even young children can understand this block diagram easily. As for Microsoft Copilot, it does not explain each steps. Instead, it only claims the overall function of this block diagram. The logic does not follow the original diagram. Besides, some parts are wrong and never mentioned in the image. For example, it misses to set one movement motor. "Turn the robot" refers to set yaw angle to 0, which is unclear and easy to cause misleading. The comments also use 'gyro' block that never appears in the diagram. This diagram explanation is unqualified because young children are not able to understand the logic of original diagram with these comments. They cannot select and build the required blocks or set the correct parameters either. LLaVA-1.5 fails again in this task. It cannot recognize this block diagram and generates some irrelevant texts. As for other two test examples, the results are similar to the first example for all of three models.

\begin{figure*}
\centering
\includegraphics[width=1\textwidth]{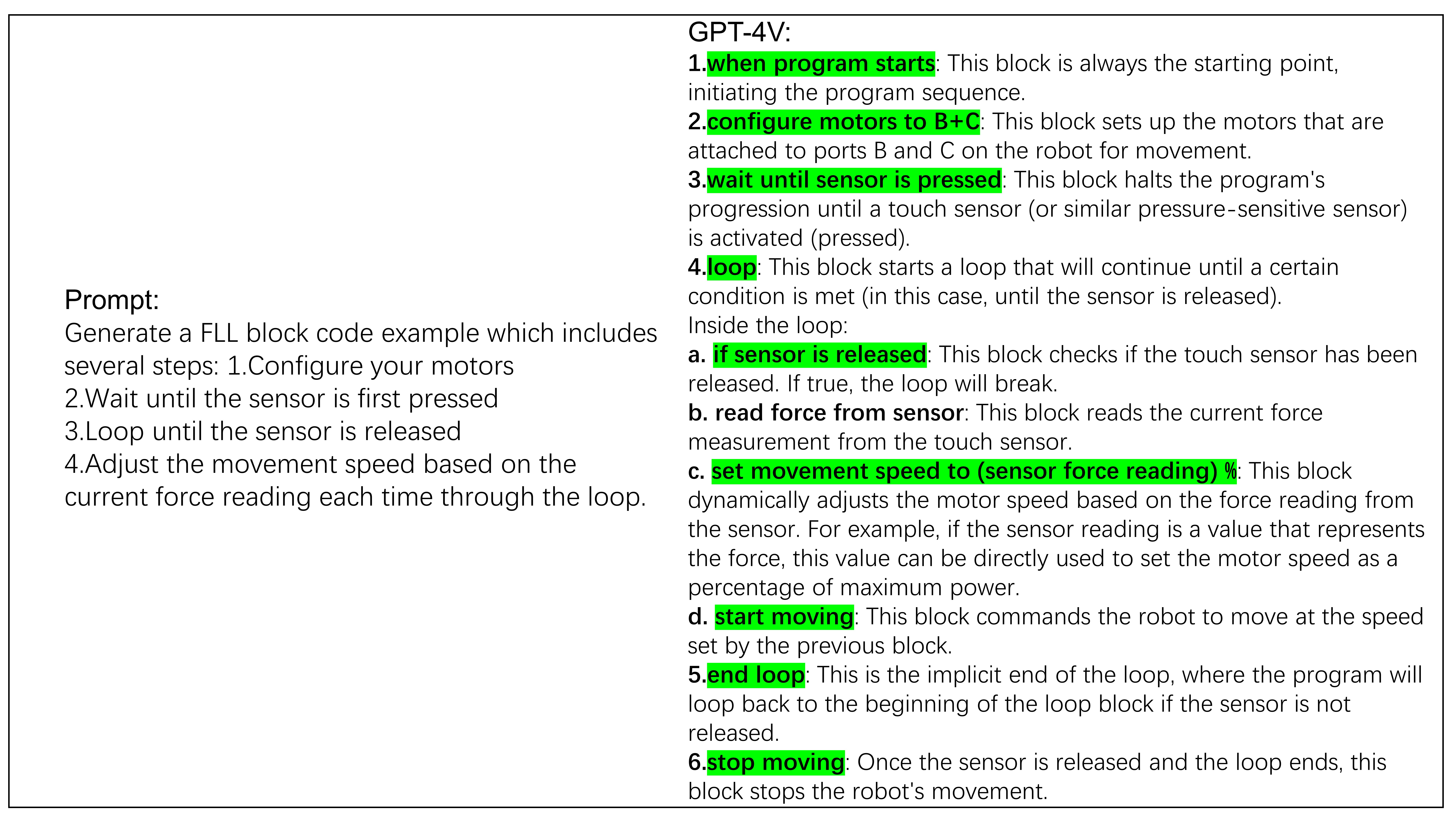}
\caption{One example for FLL block diagram text pseudocode generation results.} 
\label{gpt4_fll_gen}
\end{figure*}

\begin{figure*}
\centering
\includegraphics[width=1\textwidth]{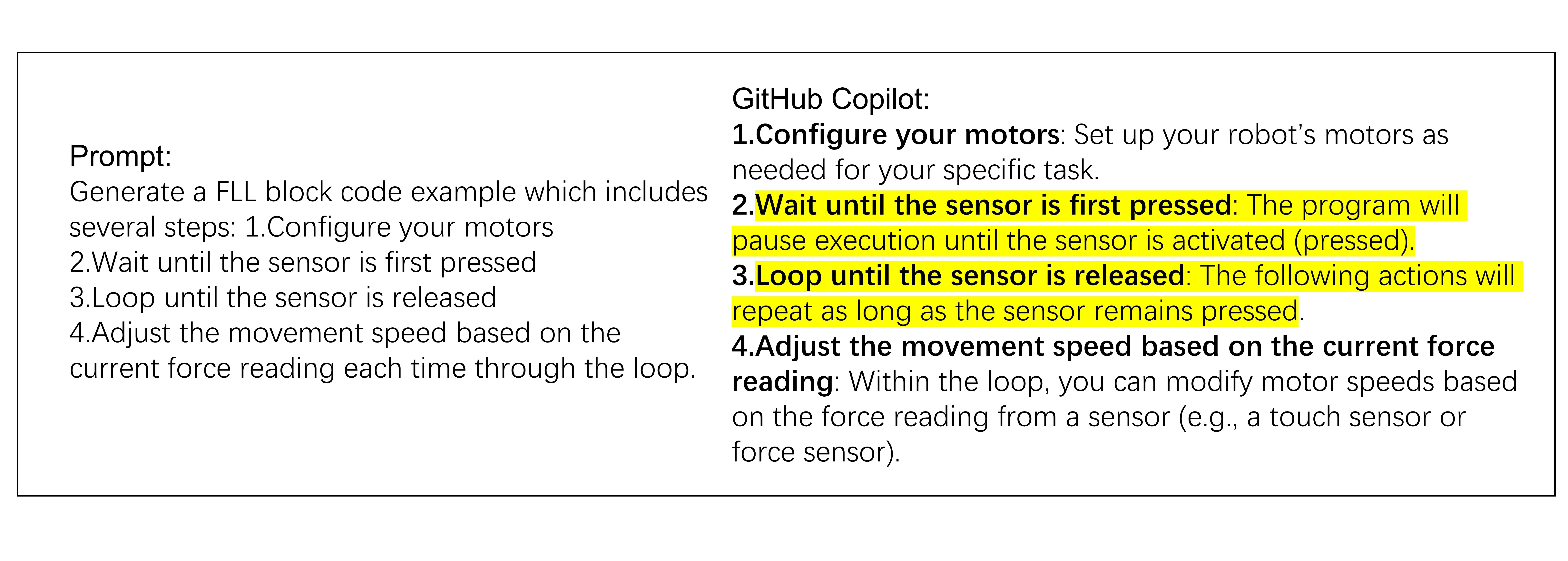}
\caption{One example for FLL block diagram text pseudo code generation results. The yellow background indicates inaccurate description. } 
\label{copilot_fll_gen}
\end{figure*}

For the final task, we apply DALL·E 3, LLaVA-1.5 and Microsoft Copilot, trying to generate direct FLL block diagrams. However, none of these models generates meaningful block diagram images with right logic. They cannot understand and include reasoning in an image. Therefore, we use GPT-4V, Code Llama and GitHub Copilot to generate names and parameter settings of each block. The results are shown in figure \ref{gpt4_fll_gen} and \ref{copilot_fll_gen}. Given the prompt with operations in each step, GPT-4V recovers each block with correct logic and sequence. The extra comments explain the corresponding functions. Although it contains one more block 'read force from sensor', young children can still follow the names of blocks to select desired blocks with correct parameters. However, the result from GitHub Copilot only repeats each instruction in input prompt with short explanations, failing to demonstrate block names as well as parameters. Code Llama cannot understand FLL block diagram and refuses to generate pseudo code again in this task.

In summary, GPT-4V outperforms than other LLMs or multimodal LLMs by finishing specific tasks with almost perfect codes and explanations. GitHub Copilot shows promise in traditional code generation and assistance but has unsatisfying performance when encountering block diagrams format. Llama series models are the worst LLMs among all the test tasks.

\section{Limitations}
%There are some limitations in our test. Firstly, we only utilize several main stream models for the test. Some other LLMs and multimodal LLMs are not covered. Secondly, we mainly focus on the robot code understanding, reasoning and generation. The test for the traditional code problems claimed in section 3.1 is not comprehensive. Although we generate code solutions for four different programming languages, we only select one problem for every difficulty level. We do not cover the entire data structures and algorithms frequently used in companies. As for FTC code, we only check the fundamental framework and logic of the code. It would be more beneficial to test the task if we can apply this technique in a real FTC competition. Due to the limited reasoning ability of multimodal LLMs, we are not able to generate direct FLL block diagrams. In the future we will keep testing new multimodal LLMs for this task. Some children might be invited to the test so that we can figure out whether any improvement is achieved during their learning process.

This study presents several limitations in the experimental methodology. Firstly, the analysis is confined to a selection of mainstream language models, excluding a comprehensive examination of other LLMs and multimodal LLMs. Secondly, our investigation predominantly centers on the capabilities of these models in understanding, reasoning, and generating robotic code, with a relatively narrow focus on traditional coding problems as delineated in Section 3.1. The methodology involves generating code solutions across four different programming languages; however, for each level of difficulty, only a singular problem is selected. This approach does not encompass the wide array of data structures and algorithms frequently employed within corporate settings. Regarding the FTC code, our evaluation is limited to the essential framework and logical constructs of the code. The practical application of these techniques in an actual FTC competition could potentially offer more insightful findings. The limited reasoning capabilities of current multimodal LLMs constrain our ability to directly generate FLL block diagrams. Future research will aim to extend the testing to emerging multimodal LLMs, enhancing the scope of this study. Additionally, involving children in the testing process may provide valuable insights into any advancements in their learning outcomes.

\section{Conclusion}
In this paper, we test several LLMs and multimodal LLMs for both traditional code generation as well as robot code understanding and generation. We find that GPT-4V outperform than other models in all of our tests while Llama series models show the worst results. However, all models suffer from the failure when generating direct block diagrams. We demonstrate the crucial trend applying LLMs and multimodal LLMs into robot education for young people. We hope our test can inspire a closer integration of advanced artificial general intelligence with code education. This approach should encompass not only traditional algorithms but also robot competitions, which demand proficiency in both software coding and robot structure building.

\bibliographystyle{splncs04}
\bibliography{mybib}

% \end{CJK}
\end{document}